# Towards accelerating physical discovery via non-interactive and interactive multi-fidelity Bayesian Optimization: Current challenges and future opportunities

Arpan Biswas[1,a], Sai Mani Prudhvi Valleti[2], Rama Vasudevan[3], Maxim Ziatdinov,[4] Sergei V. Kalinin[4,5,b]

[1] University of Tennessee-Oak Ridge Innovation Institute, Knoxville, USA
[2] Bredesen Center for Interdisciplinary Research, University of Tennessee, Knoxville, USA
[3] Center for Nanophase Materials Sciences, Oak Ridge National Laboratory, USA
[4] Physical Sciences Division, Pacific Northwest National Laboratory, Richland, WA 99352 USA

[5] Department of Materials Science and Engineering, University of Tennessee, Knoxville, USA

Both computational and experimental material discovery bring forth the challenge of exploring multidimensional and often non-differentiable parameter spaces, such as phase diagrams of Hamiltonians with multiple interactions, composition spaces of combinatorial libraries, processing spaces, and molecular embedding spaces. Often these systems are expensive or time-consuming to evaluate a single instance, and hence classical approaches based on exhaustive grid or random search are too data intensive. This resulted in strong interest towards active learning methods such as Bayesian optimization (BO) where the adaptive exploration occurs based on human learning (discovery) objective. However, classical BO is based on a predefined optimization target, and policies balancing exploration and exploitation are purely data driven. In practical settings, the domain expert can pose prior knowledge on the system in form of partially known physics laws and often varies exploration policies during the experiment. Here, we explore interactive workflows building on multi-fidelity BO (MFBO), starting with classical (data-driven) MFBO, then structured (physics-driven) sMFBO, and extending it to allow human in the loop interactive iMFBO workflows for adaptive and domain expert aligned exploration. These approaches are demonstrated over highly non-smooth multi-fidelity simulation data generated from an Ising model, considering spin-spin interaction as parameter space, lattice sizes as fidelity spaces, and the objective as maximizing heat capacity. Detailed analysis and comparison show the impact of physics knowledge injection and on-the-fly human decisions for improved exploration, current challenges, and potential opportunities for algorithm development with combining data, physics and real time human decisions. The associated notebooks allow to reproduce the reported analyses and apply them to other systems (https://github.com/arpanbiswas52/iMFBO_Ising).

**Keywords**: Human intervened autonomous exploration, Multi-fidelity Bayesian optimization, Structured Gaussian Process, Ising Model

[a] abiswas5@utk.edu

[b] sergei2@utk.edu

The exploration of expansive parameter spaces stands as a fundamental endeavor in various scientific domains. This exploration is pivotal in fields like statistical physics, particularly in the context of Ising[1–4], Potts[5–7], and Heisenberg models[8–10], as well as in the study of quantum Hamiltonians[8,10–17] and the computational construction of phase diagrams within the CALPHAD approach[18–21]. Parallel challenges arise in experimental settings, where the comprehensive investigation of combinatorial libraries encompassing oxides, metals, and assorted compounds is crucial[22–27]. The automation of material synthesis[28–32], employing techniques like pipetting and microfluidic robotics, along with the optimization of complex electrolytes and solvents mixtures, further exemplifies the breadth and importance of this task.

The classical paradigm for exploring these spaces is associated with forms of grid-based and random sampling[33,34]. However, these studies are associated with time consuming or expensive steps, and necessitate the development of more efficient approaches for the exploration of parameter spaces, typically implemented in the form of active learning. In this approach, the result of the experiment at several initial points in the parameter space are used to establish the location for the next experiment, and the process is repeated iteratively in pursuance of a selected optimization target. These may include the discovery of the target function behavior over the full parameter space, or optimization of the target function. In the context of human-driven optimization, very often the exploration task balances the data driven strategies with discovery of physical laws governing the behavior of the functionality of interest within the search space. When available, this physical knowledge significantly accelerates materials optimization.

The classical paradigm for active learning is Bayesian Optimization (BO) based on the Gaussian Processes (GP) model. In this approach, the GP is used to construct the surrogate function representing likely behavior of the target function over the parameter space, along with its uncertainty. The predicted function and its uncertainty can be combined to formulate the acquisition function, balancing the exploration and exploitation. A number of excellent reviews on Bayesian Optimization are available[35,36], and it is now implemented in a broad range of Python libraries including BOtorch (Pytorch)[37], Gpax (Jax)[38].

However, the significant limitation of classical GP is its purely data-driven nature. In other words, the surrogate function is typically constructed only from previous data. While convenient and justified in certain scenarios, in practical settings multiple sources of information about the target function are often available. These can include the availability of a certain proxy measurements that are correlated, but non-equivalent, to the target function. This proxy measurement can be known in advance or be also available dynamically, with lower measurement cost. In the latter case, the exploration of the parameter spaces includes developing strategies for balancing the low-cost (low-fidelity) and high-cost (high-fidelity) measurements. In the former case, the proxy measurement or the low fidelity data can be used to optimize exploration strategy for the expensive (high-fidelity) function. These strategically balanced low and high-fidelity data can be learned jointly through a multi-fidelity GP, to estimate the high-fidelity function of interest (ground truth), thereby reducing the cost/number of high-fidelity measurements needed to accelerate scientific discoveries, unlike in single fidelity GP[39].

The second type of knowledge available is that of partial knowledge of the physics of the system. For statistical physics this will include information on the number and potentially character of phase transitions, asymptotic behavior of the target function, etc. For exploration of phase diagrams, there can be physical models for the concentration dependence of the parameter of interest. These cases are highly domain specific, and incorporation of the physics – based mean function as a probabilistic model has been demonstrated recently for the perovskite discovery[40] and automated experiment in scanning probe microscopy[41].

In many cases multiple types of knowledge are available, necessitating strategies to balance them. In a classical BO workflow, the acquisition function is constructed prior to the optimization campaign. If necessary, the corresponding hyper parameters are optimized using model systems or pre-acquired data. This approach, while valid, is limited in many experimental scenarios and also contrasts the intended goal of automated experiments. At the same time, human-based exploration workflows often switch between different forms of exploration, alternating between data driven and physics-based discovery, and often transitioning from more exploratory policies in the beginning towards greedy optimization policies by the end of the experimental campaign.

Here, we propose to use the interactive human in the loop workflow for physics discovery based on multi-fidelity Gaussian Process with the structured probabilistic model as a mean function. We explore the limitation of simple multi-fidelity approaches as applied to function optimization. We develop the corresponding workflow and demonstrate it for the model function, and subsequently apply it for the exploration of the model Ising model. The proposed approach is universal and can be extended for exploration of multicomponent parameter spaces both in theoretical and experimental settings.

## 1. Methodology

In this section, we review the detailed architecture of standard multi-fidelity Bayesian optimization (MFBO), and then expand it to the proposed structured multi-fidelity Bayesian optimization (sMFBO). Bayesian optimization (BO)[36,42,43] is an active learning approach which is designed to provide a derivative free optimization task of an expensive (time/resource) and black-box problem, where the functional representation between the input (control parameter) and the output (experimental results) is unknown and the optimal solution can be found with minimal experiments. Here, starting with experimental results at few random locations, a computationally cheaper surrogate Gaussian process (GP) model[44–46] that replaces the unknown function is fitted to predict the outcome of the non-explored locations in the parameter space. Then the future experiments are suggested through maximizing the acquisition function[47–50], driven by the human preferred static objective. This iterative process of GP updating with new experiments and the adaptive suggestion of the acquisition function for future experiments converges to find the optimal control parameters. This type of autonomous workflow with Bayesian optimization has been widely used in recent studies to adaptively explore control parameter spaces of physical models[51–55] and to develop autonomous platforms towards accelerating chemical[56,57] and material design[58–60]. Recently, interactive Bayesian optimization frameworks through minor human intervention (human in the loop) proved to have better material processing[61] and microscope experimental steering[62–65].

Expanding BO to multi-fidelity BO or MFBO, allows the surrogate GP model to learn or train from multi-fidelity data. In the simplest case, multi-fidelity data is comprised of high-fidelity data and low-fidelity data, where the high-fidelity data are generated from time consuming (expensive) experiments or simulations and the low-fidelity data are generated from computationally cheaper experiments or simulations.

Depending on the context, one can think of high-fidelity data are the real data from the actual system that the operator wants to study and interpret valuable information, and the low-fidelity data are the approximated data from the proxy system. The aim of the MFBO is to further reduce the cost of expensive experiments or simulations where a single GP trains from multi-fidelity data and predicts the outcome of the non-explored locations in both the fidelity space. Unlike the single fidelity acquisition function in BO provide guidance on only future high-fidelity experiments, the multi-fidelity acquisition function in MFBO provides joint guidance of future experiments and the fidelity level (high or low) of the experiments. Thus, MFBO facilities in reducing the redundant expensive experiments in potential non-optimal regions over the parameter search space through gathering knowledge from low-fidelity data.

In recent years, MFBO has come to the limelight and has been implemented in accelerating discoveries in the domain of material science. A few notable examples include the application of MFBO to high throughput materials screening[66], to search a large candidate set of covalent organic frameworks for largest Xe/Kr equilibrium adsorptive selectivity[67], design and optimization of soft and biological materials[68], and multiscale design of a piezoelectric transducer[69]. MFBO can be utilized to learn from combining multi-fidelity atomistic materials simulations such as density functional theory and classical interatomic potential calculations[70], or to learn from combining data generated from experiments and atomistic simulations[71]. To the best of our knowledge, MFBO has not been integrated to explore the large space of statistical or quantum physics models (eg. Ising, Heisenberg, quantum Hamiltonians), and therefore, our study aims to extend the application to such physical models, identify current bottlenecks and future opportunities to gain deeper insights and accelerate physical discoveries, which won't be possible through exhaustive, grid-based, or even with single fidelity BO. To illustrate the architecture of the standard MFBO, below **Table 1** demonstrates the algorithm. It is to be noted the algorithm is demonstrated for two fidelity levels, however it can be easily extended for >2 fidelity levels.

**Table 1: Algorithm: Multi-fidelity Bayesian optimization (MFBO).**

1. **Initialization:** State maximum BO iteration, $M$. Randomly select $j$ samples with randomly chosen fidelity levels $f = 0\ (low)\ or\ 1\ (high)$. Build the multi-fidelity input data matrix $\boldsymbol{X^F} = \{\boldsymbol{x_1}, f_1;\ \boldsymbol{x_2}, f_2, \ldots \boldsymbol{x_j}, f_j\}$, where the last column represents the fidelity level and the other columns represents the input parameter values. Calculate the output $\boldsymbol{Y^F} = \{y_{1,f_1}, y_{2,f_2}, \ldots y_{j,f_j}\}$. Build the multi-fidelity training data, $\boldsymbol{D^F} = \{\boldsymbol{X^F}, \boldsymbol{Y^F}\}$.

**Start BO.** Set $k = 1$. For $k \leq M$

2. **Multi-fidelity Surrogate Modelling**: Develop or update MFGP models, given the training data, as $\Delta^F(D^F{}_k)$. Here, we design the multi-fidelity kernel function adapting from [35] and is implemented in *GPax* Python library package[72].

   a. Here, the mean function of the $\Delta^F$ is 0.
   b. The covariance or **multi-fidelity kernel function** of $\Delta^F$ is defined in the Method Section as per eqs. 5-8.

   c. Optimize the hyper-parameters of kernel functions of $\Delta^F$. Here, we optimize through Bayesian analysis (posterior estimation) in Monte Carlo Markov Chain (MCMC) [73] by maximizing the log-likelihood (learned from the data).

3. **Posterior Predictions**: Given the surrogate model $\Delta^F$, compute high and low fidelities posterior means and variances for the unexplored locations, $\overline{\overline{X_k^F}}$, over the parameter space as $\mu_{HF}(Y(\overline{\overline{X_k^F}})|\Delta^F$, $\mu_{lf}(Y(\overline{\overline{X_k^F}})|\Delta^F$ and $\sigma_{HF}^2(Y(\overline{\overline{X_k^F}})|\Delta^F, \sigma_{lf}^2(Y(\overline{\overline{X_k^F}})|\Delta^F$ respectively.

4. **Multi-fidelity Acquisition function:** Compute and maximize the multi-fidelity acquisition function (MFAF), $\max_X U^F(.|\Delta^F)$ to select 1) next best location and 2) fidelity level as, $\{x_{k+1}, f_{k+1}\}$ for evaluations. Here, we design the MFAF adapting from [74], following the Expected Improvement based acquisition function, and is described in Method Section as per eqs. 9-15.

5. **Function evaluations:** Depending on the fidelity level, $f_{k+1}$, as suggested via the optimization task in Step 4, evaluation is conducted at location $x_{k+1}$, and the output is generated as $y_{k+1,f_{k+1}}$.

6. **Augmentation:** Augment data, $D_{k+1}^F = [D_k^F; \{x_{k+1}, f_{k+1}\}, y_{k+1,f_{k+1}}\}$. Repeat Step 2 to 6 until convergence.

Fig. 1 shows a simple example of the MFBO exploration over two synthetic 1D test problem. The formulation of the test functions is provided in Supplementary Materials. The first test function (fig. 1a) in both fidelities are continuous everywhere which aligns with the assumption for fitting the surrogate model in BO.

We started with 10 randomly generated samples and then autonomously explored for 15 iterations to learn the peak of the function (region of interest). In the starting samples (fig. 1a), we have 7 low-fidelity observations and 3 high-fidelity observations. Also, an interesting point to note, the starting samples are all far away from the region of interest (peak of the high-fidelity function) which we aim to reach via autonomous exploration over MFBO. Here, the cost ratio $C = \frac{5}{4}$ (following eq. 14 in Method Section). As the MFBO iterates (fig. 1b), it exploits with high fidelity observations in the peak of the function, which was sampled during the initialization, presumed as the potential region of interest as it still did not explore near the actual region of interest. We can see only after iteration 12 (fig. 1c), the MFBO explores and samples a low-fidelity observation near the region of interest, which suggests for the high-fidelity exploitation to uncover the potential region of interest (refer to the subfigure in fig. 1c). Then, at iteration 15 (fig. 1d), we can see the MFBO finds the true optimal solution with high-fidelity exploitation. Overall, the MFBO suggested 6 high-fidelity and 9 low-fidelity evaluations, with the trend to explore to learn the unknown areas with low-fidelity evaluations and exploit the potential region of interest with high-fidelity evaluations.

The second test function (fig. 1e) is more challenging which has discontinuity (at $x = 7.5$) on the high-fidelity level but is continuous in the low-fidelity (approximated) level. In other words, unlike the first problem, here the low-fidelity function is partially correct where it does not have information of the discontinuity. This is very relevant with the real-world problems where the low-fidelity approximated system can miss the actual physical properties of the system due to partial knowledge of the physics. However, the goal would be still the MFBO to intelligently explore and provide the discovery of the true knowledge of the system. Here, the cost ratio $C = 2$.

In fig. (1f), after 15 iterations, the MFBO provides the optimal solution which slightly deviates from the true optimal solution. Interestingly, it also suggests exploiting with low-fidelity observations at the region of interest due to the lack of the knowledge of the discontinuous peak in the ground truth (high-fidelity function). In this case, the MFBO suggests 4 high-fidelity evaluations and 11 low-fidelity evaluations. The MFBO underfits with continuity assumption of the low-fidelity models, therefore, sways the algorithm towards exploiting the low-fidelity function peak (not the actual human preferred region of interest). This is because the connection between low and high fidelity is symmetric. Also, due to the partially incorrect low-fidelity model, the measure of uncertainty is higher in this case and the MFGP provides weakly aligned high and low fidelities mean plots with respective to the actual functions.

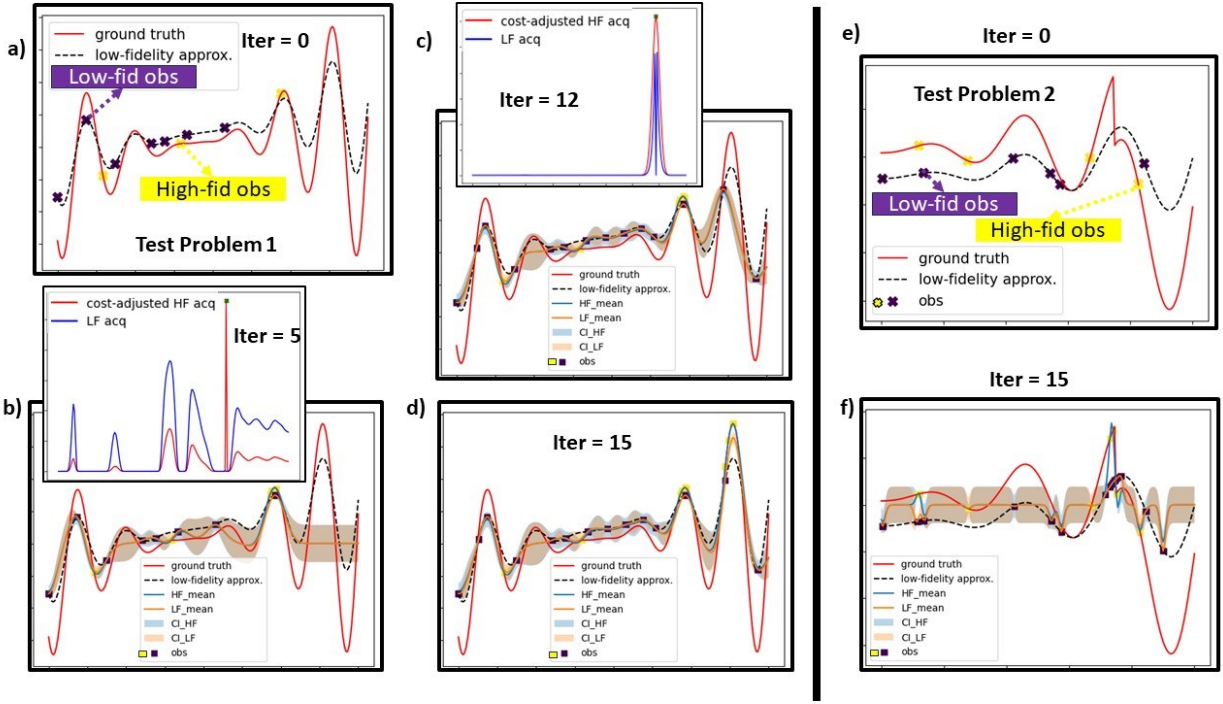

**Fig 1.** Multi-fidelity BO exploration over test problems. Fig 1a – Fig. 1d are the analysis over test function 1 which is continuous everywhere. Fig 1e– Fig. 1f are the analysis over test function 2 which has discontinuity. The red solid line is the actual ground truth or high-fidelity function, and the black dotted line is the approximated low-fidelity function. The yellow and the black squares in the legend denote the high-fidelity and low-fidelity observations, as autonomously sampled from MFBO. The subfigure within Fig. 1b and 1c are plots of the high-fidelity (red) and low-fidelity (blue) acquisition values over the parameter X, at the specified BO iteration. It is to be noted we choose the value and the fidelity of X, as $x_{k+1}, f_{k+1}$, based on the location of the highest peak between the acquisition graphs.

The analysis from these motivating examples highlights two limitations of standard MFGP in convergence towards accelerating discoveries (reducing time/resources) in complex situations such as 1) when the true region of interest is hard to find and 2) when the low-fidelity models are only partial-informative to the true physics of the system. The reason behind such limitations is generally the standard MFBO is only data-driven and does the poses any known physical behavior of the system which the domain experts have. Also, the standard MFBO, like any benchmark AI tools, has a rigid architecture where it does not have the flexibility for "on the fly" adjustment with human intervention to steer toward the objective of the domain experts. The focus of this paper is to address these situations and thereby aim to provide an improved adaptive and interactive search of BO in multi-fidelity problems, combining data-driven approach with physics-driven and minor human intervention.

## Human-in-the loop based Automated Experiment (hAE) Workflow: Interactive Multi-fidelity Bayesian Optimization

Here, we present the interactive MFBO (iMFBO) where the domain expert, starting with a problem objective and BO search policy (parameter space, surrogate model, acquisition function, convergence criteria etc.), has the flexibility to provide on the fly policy adjustment based on the monitoring of the current BO progress and thereby improve experimental steering towards alignment with the objective. iMFBO contains the option to avail the structured MFBO (sMFBO) where the MFGP is injected with known physical behavior of the system and minor human intervention to minimize any on the fly AI misalignment. **Fig 2.** provides the architecture of iMFBO where the red highlights show the research contribution. Here, following Table 1, the modification is done on Step 2a where unlike the zero mean of the fitted standard MFGP model, the structured MFGP has a local mean function probabilistic model $M_f$ which provides the flexibility to insert the domain knowledge (local physical behavior) of the high-fidelity models. It is to be noted this type of domain knowledge injection process has been configured earlier in BO[75], here for the first time to our knowledge, we have extended the approach to the multi-fidelity setting. Also, we added Step 7 in Table 1 to incorporate the policy adjustment with human interactions. Table 2 provides an example on human interacted policy setup; however, it is evident to mention that any interactive policy can be designed in this similar approach.

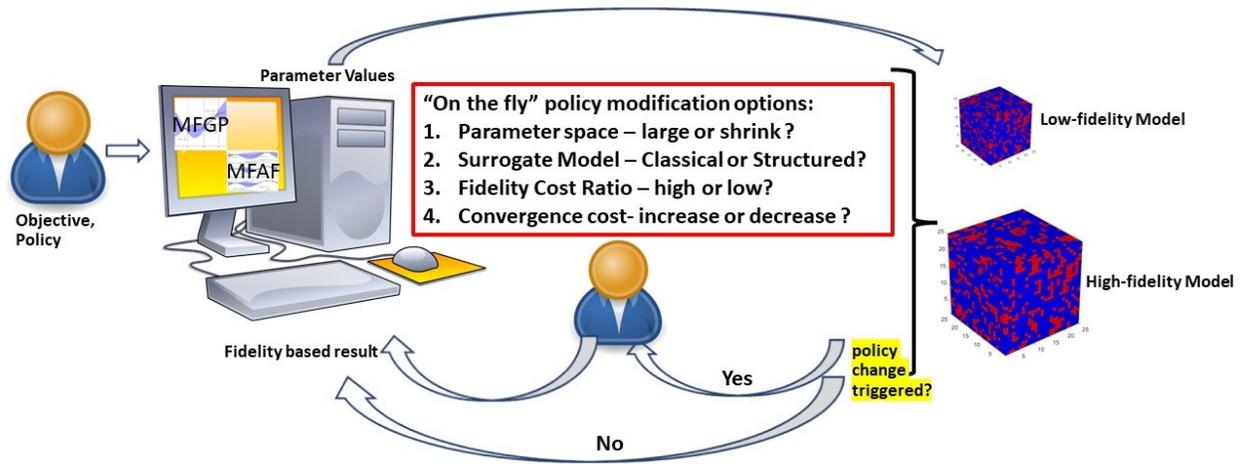

**Fig 2.** Design of the proposed interactive multi-fidelity Bayesian optimization (iMFBO), a human in the loop automated experiment approach (hAE). In this structure, the domain expert (human) starts with a problem objective and the BO search policy (parameter space, surrogate model, fidelity cost ratio in acquisition function, convergence criteria etc.) and then through monitoring the BO progress (alignment with domain expert objective) has the flexibility to provide on the fly policy modification. During the course of the exploration, the domain experts can turn down policy change, or decide to select either single or multiple option choices as per real time knowledge gain. Here, with on-the-fly selection of structured surrogate model, the existing physics knowledge derived from the high-fidelity model can be injected into the MFGP mean function and the steering of the high vs low fidelity sampling during real time exploration can be done through adjusting fidelity cost ratio.

**Table 2: Algorithm: Extension of MFBO to iMFBO. Here, we have provided the added Step 7 to the standard MFBO in Table 1.**

7. **Interactive Policy:**
    a. Calculate the current best solution as $y(x^+)$ among all the fidelity spaces.
    b. *Interactive Step:* User interaction after $i < k < M$ iterations without any current solution improvement, i.e., $y_{k-i,f_{k-i}}, \ldots, y_{k+1,f_{k+1}} \leq y(x^+)$. User gets a prompt question if the policy needs to change. User has option to choose, Yes or No. If "No", proceed to Step 2 (Table 1) with the current policy. If "Yes", proceed to Step 7c.
    c. *Interactive Step:* User gets options of the policies that can be changed as "parameter space", 2) "surrogate model", 3) "acquisition function", 4) "convergence criteria". User can select multiple options. Based on the options selected, all or some steps (d)-(g) will be followed.
    d. *Interactive Step:* If user select "parameter space", user gets a prompt message to provide the new values of minimum and maximum input values for each dimension. Given the user input, a new matrix of unexplored locations, $\overline{X^F_{k_{new}}}$, will be generated.
    e. *Interactive Step:* If user select "surrogate model", user gets a prompt message to choose a different surrogate model. Here, we have two options of surrogate model as standard MFGP (only data driven) and sMFGP (physics and data driven). It is to be noted, that any number of surrogate models can be ensembled which can be controlled with such "on the fly" decisions.
    f. *Interactive Step:* If user selects "acquisition function", user gets a prompt message to provide new cost ratio, $C_{new}$, between the fidelities.
    g. *Interactive Step:* If user select "convergence criteria", user gets a prompt message to provide new value of total iterations, $M_{new} > k$, such that remaining $M_{new} - k$ iterations will be conducted with the new policy.
    h. Repeat Step 2-7 until convergence.

---

To validate the proposed iMFBO, we consider the test problem 2. Here in the physics driven sMFGP, we inject the knowledge of discontinuity of the high-fidelity function through developing the MFGP mean function model as below.

$$M_f(x, a, b, c) = \begin{cases} f(x) - a \text{ if } x < c \\ f(x) - b \text{ if } x \geq c \end{cases} \tag{1}$$

$$f(x) = f_1(x) = -(x+1)^2 + \frac{\sin(2x+2)}{5} + 1 + \frac{x}{3} \tag{2}$$

$$f(x) = f_2(x) = x^2 \tag{3}$$

where a, b and c are unknown and the posterior estimation are computed through the Bayesian analysis in MCMC along with the hyperparameter estimation of the MFGP (refer to Table 1, Step 2c). The prior distributions are considered as $p(a) = N(0,1)$, $p(b) = N(15,2)$, $p(c) =$ $Unif[5, 10]$. In other words, the knowledge of the presence of the discontinuity over the high-fidelity function is injected into MFBO, however, the location of the discontinuity is unknown with some prior belief. This is again true for most physics model where we generally know partial information of some behavior of the system but are unfamiliar with the exact functional mapping with the control parameters. Also, we tested with two functional forms of $f(x)$ where $f_1(x)$ has the correct prior knowledge and $f_2(x)$ has the incorrect prior knowledge of the true function. We started with the same 10 samples and the cost ratio $C = 2$.

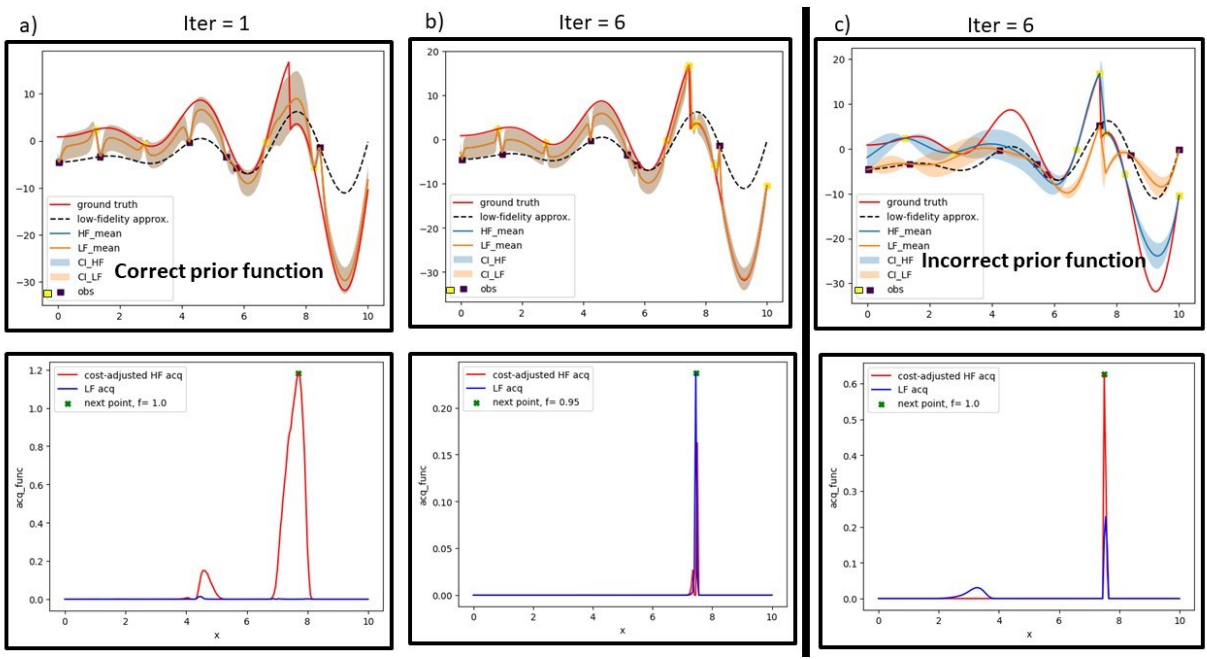

**Fig 3.** Non-interactive structured Multi-fidelity BO (sMFBO) exploration over 1D test function where high-fidelity function poses discontinuity and the low fidelity function is continuous everywhere. Fig 3a-b are the example where correct prior functional form is injected (refer to eq. 2) and Fig 3c is the example where incorrect prior functional form is injected (refer to eq. 3). The red solid line is the actual ground truth or high-fidelity function, and the black dotted line is the approximated low-fidelity function. The yellow and the black squares in the legend denote the high-fidelity and low-fidelity observations, as autonomously sampled from sMFBO. The bottom images of the figures are plots of the high-fidelity (red) and low-fidelity (blue) acquisition values over the parameter X, at the specified BO iteration.

**Fig. 3** shows the significant improvement of the overall exploration of non-interactive sMFBO and iMFBO than the standard MFBO. We can easily see at iteration 1 (fig. 3a) when the samples are far away from the actual region of interest, the non-interactive structured sMFGP learns the potential region of interest via the injected knowledge and drives the multi-fidelity acquisition function to exploit with the high-fidelity observations. After only 6 iterations (fig. 3b), the sMFBO able to converge to the human-preferred region of interest and the learned sMFGP accurately predicts the discontinuity, which the standard MFGP failed to do with more than 2x iterations. Interestingly, we observe similar performance even with providing incorrect prior functional forms where the incorrect prior knowledge has been neglected with adaptively selected multi-fidelity observations (fig. 3c). However, as sMFBO continues to iterate for 15 iterations (set convergence policy), we see the non-interactive model still provide redundant low-fidelity evaluations near the discontinuity region which has been already learned with multi-fidelity observations (refer to figure S1 in Supplementary Materials). In real cases, after the discontinuity is learned, the domain expert will tend to avoid those regions, being a very sensitive and non-robust region. Thus, in the iMFBO, the operator readjusts the policy for the remaining iterations for better learning of the function with a new region of interest with selecting options 1 - 4 in Step 7c. Here, shown in **Fig. 4**, we start the BO with same policy as in fig. 3c with considering the incorrect prior functional forms (eq. 3). As mentioned through monitoring the progress and learning the discontinuity, the operator changes the policy to avoid the discontinuous area, considers standard MFGP (assuming we don't have any new knowledge over the new parameter space), and reduces the cost ratio to drive more high-fidelity observations. In other words, we proceed to find the region of interest (maximum function values) between $x = 0$ to $x = 6$ (avoiding the learned discontinuous region) with standard MFGP once the discontinuity is found over the high-fidelity function through sMFGP. At iteration 12, another policy change is triggered due to no improvement of current best solution in last 5 iterations ($k = 5$ in Step 7b). Here, with the current state, the operator instead of meeting convergence, aims for a more curiosity driven search due to the region of $x = [4, 5]$ having higher low-fidelity observations. Therefore, another policy change is made on the fly into the iMFBO by further reducing the cost ratio and increasing the limit of BO iterations. As the operator speculated through low fidelity observations, the true high fidelity optimal solution is found at iteration 16 (fig. 4c) and we attain the true convergence in iteration 22 due to negligible acquisition function value (fig. 4d).

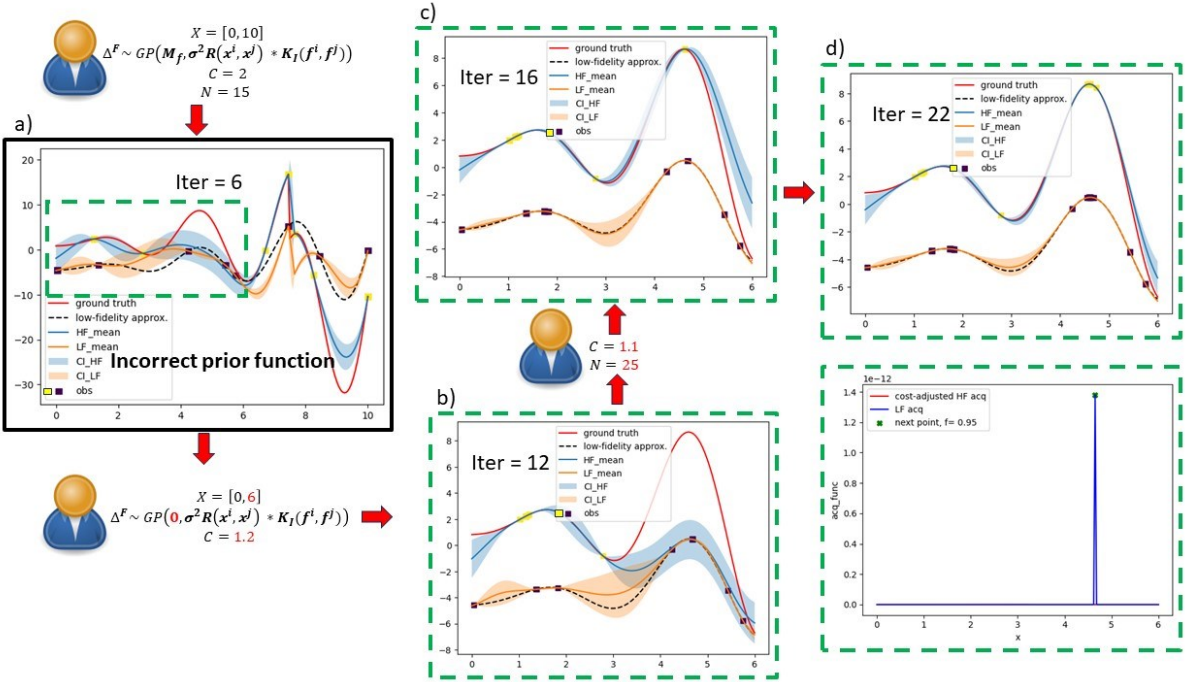

**Fig 4.** Interactive Multi-fidelity BO (iMFBO) exploration over 1D test function where high-fidelity function poses discontinuity and the low fidelity function is continuous everywhere. The red solid line is the actual ground truth or high-fidelity function, and the black dotted line is the approximated low-fidelity function. The yellow and the black squares in the legend denote the high-fidelity and low-fidelity observations, as autonomously sampled from iMFBO. In fig. 4a, the incorrect prior functional form is injected into the sMFGP (refer to eq. 3). The green dotted box in fig. 4a shows the operator's (human in the loop) on the fly policy on region of interest as highlighted in the following figs 4b-d. The bottom images of the fig. 4d is the plot of the acquisition values over the parameter X, at the specified BO iteration.

## 2. Results

Following the analysis from the benchmark problems, in this section, we conduct the analysis of such non-interactive (MFBO and sMFBO) and interactive search with iMFBO to real case studies over a statistical physics based Ising Model. Here, we considered two versions of Ising model. The first version is modeled on a two-dimensional square lattice of size $N \times N$, where $N$ represents the number of spins in one dimension. Each spin, taking a value of either +1 or -1, interacts with its nearest neighbors, encapsulating the fundamental aspects of magnetic systems. The primary input parameters for our model are the interaction strength matrix between adjacent spins and the reduced temperature $T$, which is dimensionless and scaled by the Boltzmann constant $k_B$. The interaction parameters matrix helps us model anisotropic interactions the long-range (next nearest neighbors and so on) interactions. The system's evolution is modeled using the Metropolis-Hastings algorithm, a variant of the Markov Chain Monte Carlo (MCMC) methods, renowned for its efficacy in simulating systems in statistical mechanics. The Metropolis-Hastings algorithm in our implementation begins with an initial configuration of spins, either randomly oriented or aligned, depending on the desired starting conditions. At each step, for each spin, the change in energy $\Delta E$ resulting from flipping this spin is computed. The decision to accept or reject the flip is governed by the Metropolis criterion: the flip is always accepted if $\Delta E < 0$; otherwise, it is accepted with a probability of $exp\left(-\Delta E/k_B T\right)$. At each Monte Carlo step, all the spins are This stochastic process ensures that the system explores a wide range of configurations, with a tendency to settle into states of lower energy at lower temperatures. Through repeated iterations, the system evolves, allowing for the study of equilibrium properties such as magnetization, specific-heat, and susceptibility as functions of temperature, providing insights into the nature of phase transitions in ferromagnetic materials.

For the second version of the Ising model, we selected a triangular lattice which is guided with anisotropic with the three nearest neighbors. In addition to the change in lattice structure, the second Ising model is modeled using Kawasaki dynamics. These dynamics allows the model to conserve the order parameter (magnetization) of the system. This is achieved by exchanging the positions of opposite spins as opposed to flipping them. An exchange is accepted based on the Metropolis criterion as described above. The initial magnetization of the system is user selected and is an input to the simulations. More details about these models can be found in our previous works[76–78].

In this Ising simulator model, we considered the 60x60 lattice size as high-fidelity model and a computational cheaper version of 20x20 lattice size as low-fidelity model. We considered the first 500 Monte Carlo (MC) steps to achieve model equilibrium and the next 500 MC steps for simulations with reduced temperature, $T = 2.7$. The control parameter is the spin-spin interaction parameter, $J$ and the output parameter is the simulated heat capacity, $H_c$. Here, the objective or the region of interest is to maximize the heat capacity which defines the potential phase transition of magnetism (different pattern of spin configurations in the lattice space). For simplicity, we only considered the interaction in each direction being equal, such as $J_x = J_y$ (for square model) and $J_x = J_y = J_z$ (for triangular model). Given the fact that it is extremely time consuming to explore exhaustively over 60x60 lattice size where each simulation takes about 8 mins approx., the true functional form of the heat capacity over the 60x60 model is unknown and is aimed in this study to explore with the proposed iMFBO. It is to be noted the cheaper 20x20 model takes each simulation time of about 30s. For the purpose of validation, we considered square model as the model of interest. We started with 10 randomly generated samples and then autonomously explores for 25 iterations to learn the maximum $H_c$. (region of interest). In the starting samples (fig. 1a), we have 6 low-fidelity observations and 4 high-fidelity observations. We define the cost ratio as the computational time between the high and low fidelity Ising model as $C = \frac{T_H}{t_l} * \Delta$ where $0 < \Delta \leq 1$ is the tweaking parameter to adjust the cost ratio. In this case study we considered $C = 8.6$ as per the ratio of the time complexities between high and low fidelity simulations. The prior distribution of the MFGP hyperparameters are $\sigma^2 \sim Unif(0.01, 1)$, $\theta_m \sim Unif(0.01, 1)$ and $\delta \sim Unif(0.01, 1)$. The simulation noise prior is estimated from the standard deviation of 20 low fidelity simulation at $J = 1.17$ which has the maximum $H_c$ in the low-fidelity space. Thus, we considered the noise prior as $\vartheta^2 \sim halfNormal(0.03)$.

### 2.1. *Exploration of Ising model with standard and structured MFBO (non-interactive)*:

Starting with the non-interactive standard and structured MFBO exploration **in Fig. 5**, we can definitely see the starting samples are far away from the potential region of interest (as per the low-fidelity model in fig. 5a. It is clearly seen from fig 5a, that the problem is non-trivial due to the following reasons: 1) the low-fidelity data are highly non-smooth, 2) the degree of non-smoothness of the data increases significantly towards the region of interest. Thus, we can see as we explore more towards the region of interest, which the MFAF is expected to do, the kernel scale and length decreases to fit the non-smooth data (ref to Supplementary **Figs. B2**). This is the reason the MFGP uncertainty increases significantly even at the low heat capacity region where the data is smoother, as the MFGP considers those regions as highly non-smooth as well (fig. 5c). To explore the Ising model with non-interactive sMFBO, we leverage the existing physics knowledge of the unimodal functional form with single peak at the transition region. Therefore, the mean model, $M_f$, can be defined as

$$M_f(x, a, b, c) = a * \exp\left(\frac{(x-b)^2}{2*c^2}\right) \qquad (4)$$

where $a, b, c$ are the height, location, and broadness of the peak of the function. The prior distributions of the unknown variables are considered as $p(a) = halfN(0.5)$, $p(b) = halfN(1.17)$, $p(c) = halfN(2)$. Here, the knowledge of the presence of the phase transition or the peak in $H_c$ over the high-fidelity function is injected into MFBO, however, the location of the discontinuity is unknown with some prior belief through low-fidelity observations. It is to be noted, for the purpose of comparison, the starting samples are similar to the analysis in earlier MFBO case (fig. 5a). Here also, as we explore more in the region of interest with high degree of non-smooth data, the kernel scale and length got reduced and the overall sMFGP fit get worsened (ref to Supplementary **Figs. B3**). This shows the fitting of several non-smooth data overpowers the injected physical information of a smooth gaussian function. Unlike the observation in standard MFBO, for structured sMFBO, we see the fit gets better with 35 autonomous multi-fidelity explorations, and the uncertainty overall gets lower (fig. 5e). Overall, the MFBO suggests 5 high-fidelity and 30 low-fidelity evaluations, while the sMFBO suggests 4 high-fidelity and 31 low-fidelity evaluations. Comparing with the ground truth as provided in fig. 5b. with the high-fidelity prediction function obtained after 35 MFBO and sMFBO iterations, from figs 5d, 5e, we can see the highest square error (SE) is around the high heat capacity region (where data is highly non-smooth) whereas the lowest SE is around the low heat capacity region (where data is smooth). The overall mean square error (MSE) is $8 \times 10^{-4}$ and $7 \times 10^{-4}$ respectively and thus shows the sMFBO provides better performance with the inclusion of the physical information of heat capacity functional form.

In **Fig 6**, we inserted another level of complexities as we considered the incorrect low-fidelity model (triangular model). Similar to the previous case, the kernel length and scale decreases as we explore the non-smooth region. Since the low fidelity function is incorrect, and with low fidelity exploration, the MFGP performs worse and tends to fit to the incorrect function and provide higher erroneous knowledge of high-fidelity function (refer to fig 6c, 6d). This resulted to a high MSE of $4 \times 10^{-3}$. Overall, the MFBO suggests only 1 high-fidelity and 34 low-fidelity evaluations. Additionally, we tested constraining the kernel length and scale with high lower limit values as 0.1, as considering the MFGP fit to ignore the non-smoothness of the data. We find the performance of the MFGP get worsen where the function fit becomes more erroneous, which is reasonable as we discard the true optimal kernel length and scale which minimizes the loss function of fitting the data. However, interestingly, for the sMFBO case (refer to fig 6e, 6f), the performance is much better with significant lower MSE of $5 \times 10^{-4}$. This could be due to inclusion of the prior knowledge of functional form as opposed to the classical version. Here, the sMFBO suggests 2 high-fidelity and 33 low-fidelity evaluations.

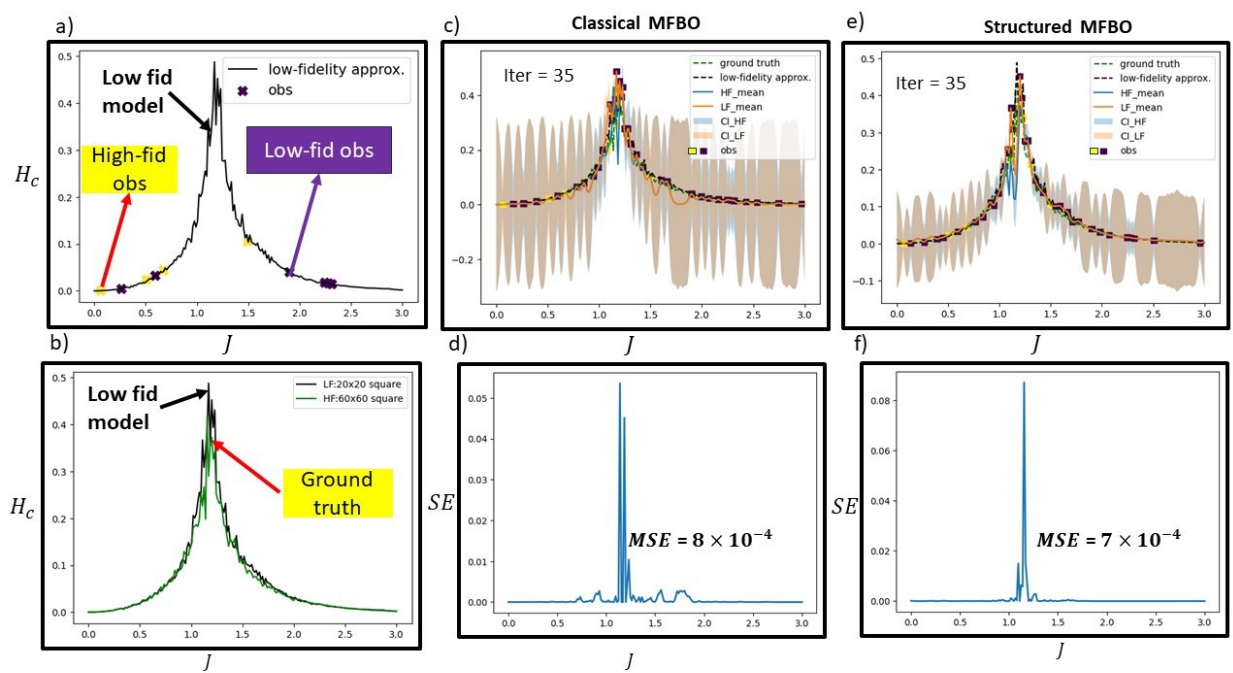

**Fig 5.** Non-interactive Multi-fidelity BO exploration over Ising models where the input is the spin-spin interaction, $J$, output is heat capacity, $H_c$, low fidelity simulation of lattice size 20 and high-fidelity simulation of lattice size 60. Here, we consider the correct model of interest (square model) in the low-fidelity space (denoted by black line) in fig (a) and the ground truth of high-fidelity function (denoted by green line) is superimposed in fig (b). Fig (c) shows the results after 35 MFBO iterations and fig (d) shows the error function between the ground truth and the prediction of high-fidelity function. Figs (e), (f) show the similar analysis for sMFBO. The yellow and the black squares in the legend of fig. (c), (e) denote the high-fidelity and low-fidelity observations, as autonomously sampled from MFBO.

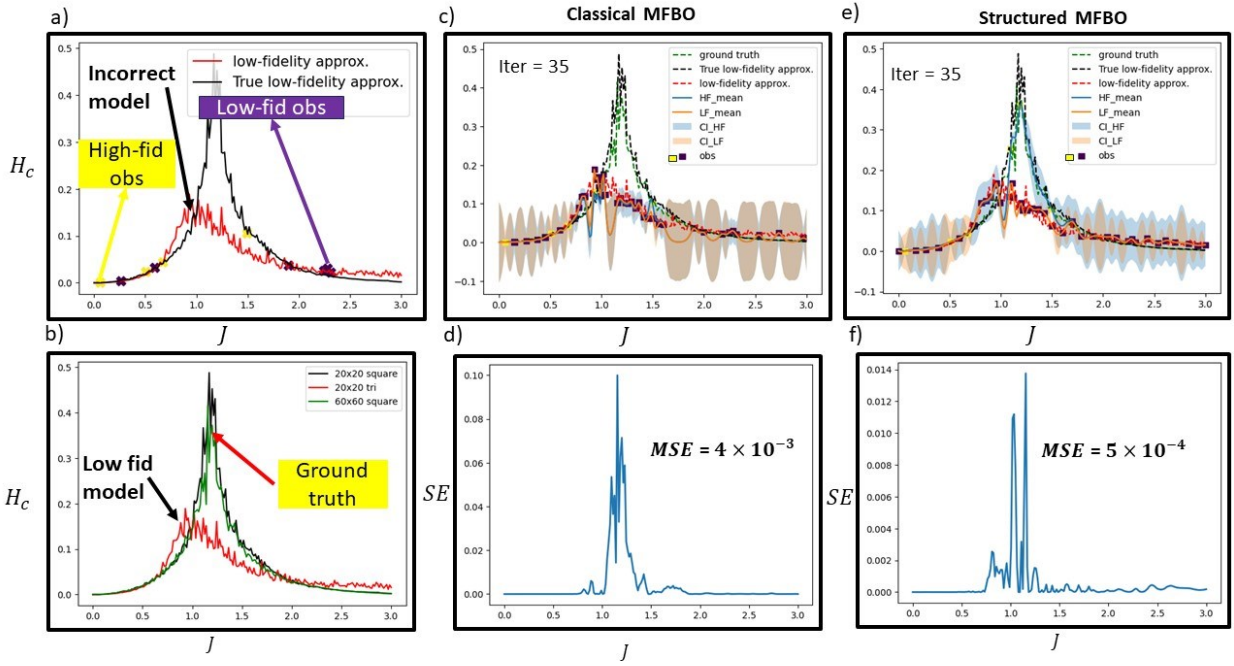

**Fig 6.** Non-interactive Multi-fidelity BO exploration over Ising models where the input, output and the fidelities of the model are same as mentioned in Fig. 5. Here, we consider the incorrect model of interest (triangular model) in the low-fidelity space (denoted by red line) in fig (a) and the ground truth of high-fidelity function (denoted by green line) is superimposed in fig (b). Fig (c) shows the results after 35 MFBO iterations and fig (d) shows the error function between the ground truth and the prediction of high-fidelity function. Figs (e), (f) show the similar analysis for sMFBO. The yellow and the black squares in the legend of figs. (b)- (e) denote the high-fidelity and low-fidelity observations, as autonomously sampled from MFBO.

### 2.2. *Exploration of Ising model with proposed interactive MFBO*:

**Fig 7 and 8** show the exploration with interactive iMFBO, considering the correct and incorrect low-fidelity models respectively. Here, we start with the same initial samples, and initialize with standard MFBO. At iteration 6 (fig. 7a), as no significant improvement can be seen from the exploration with standard MFGP, a user interaction is triggered if the current policy needs to be changed. Here, the user transit the standard MFGP to sMFGP on the fly based on monitoring the exploration. Next, after iteration 25 (fig. 7b), we can see the sMFGP has decently fit the low and high-fidelity function and therefore, another policy is changed on the fly where we force stop the explorations with the final high-fidelity exploration. Thus, we can see 4 high fidelity and 22 low-fidelity explorations with significantly lower MSE of $3 \times 10^{-4}$ (refer to **Fig. 9a**). When considering the incorrect low-fidelity model, following the same policy strategy, we can see the performance is not good as in the previous case. This could be due to the involvement of classical MFBO in early iterations which fail to learn the high-fidelity function with incorrect low fidelity data as we have seen for the non-interactive MFBO case. However, the performance is better than the standalone MFBO as we switched to sMFBO in the later part of the explorations. This showcases the importance of physical knowledge inclusion, even if the low-fidelity model is significantly erroneous. Here, we can see 6 high fidelity and 29 low-fidelity explorations with MSE of $1 \times 10^{-3}$ (refer to **Fig. 9b**).

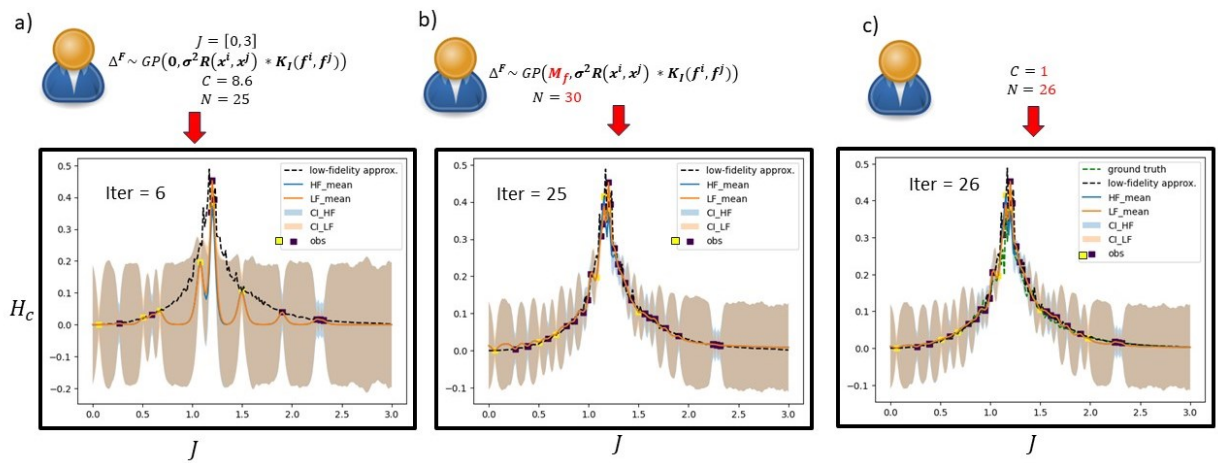

**Fig 7.** Interactive Multi-fidelity BO (iMFBO) exploration over Ising models where the input, output and the fidelities of the model are same as mentioned in Fig. 5. Here, we consider the correct model of interest (square model) in the low-fidelity space (denoted by black dotted line). The yellow and the black squares in the legend of the figures denote the high-fidelity and low-fidelity observations, as sampled from iMFBO. Here, we start the policy with standard MFBO and cost ratio defined by time complexities (fig. a), then changed the policy to structured MFBO (fig. b), finally changed the policy to relax high-fidelity explorations (fig. c).

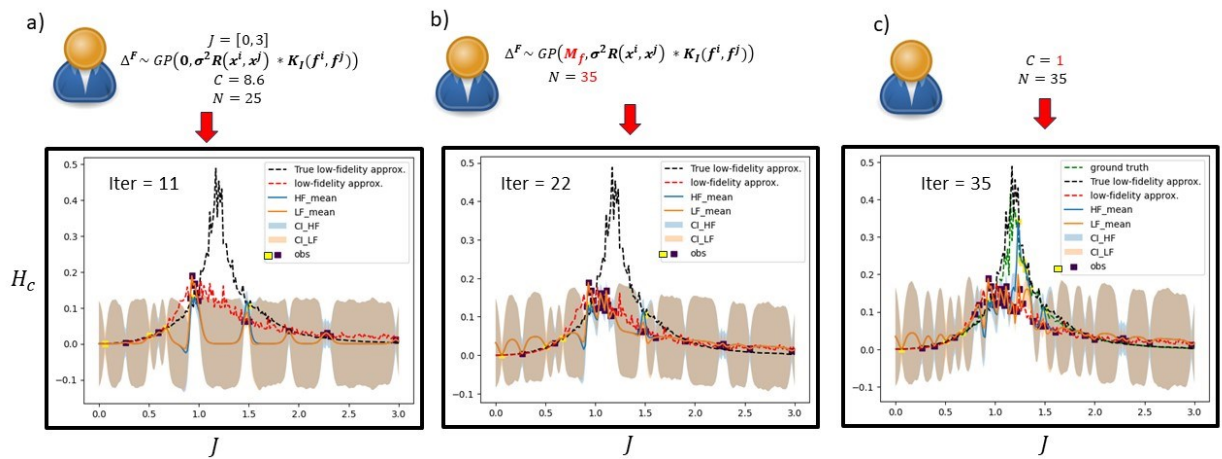

**Fig 8.** Interactive Multi-fidelity BO (iMFBO) exploration over Ising models where the input, output and the fidelities of the model are same as mentioned in Fig. 5. Here, we consider the incorrect model of interest (triangular model) in the low-fidelity space (denoted by red dotted line). The yellow and the black squares in the legend of the figures denote the high-fidelity and low-fidelity observations, as sampled from iMFBO. The strategy of the policy change is similar to mentioned in fig. 7.

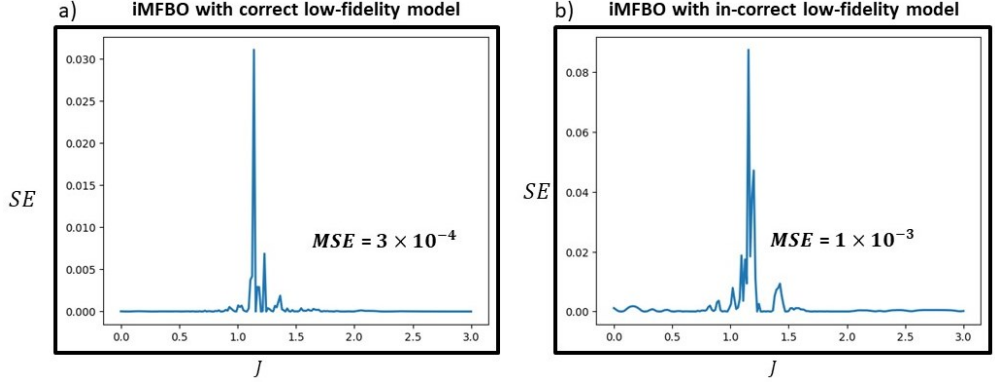

**Fig 9.** Square errors from Interactive Multi-fidelity BO (iMFBO) exploration over Ising models for the respective analysis in figs. 7, 8 respectively. Here, the input, output and the fidelities of the model are same as mentioned in Fig. 5.

To further validate the need of multi-fidelity BO, we compare the above analysis with single fidelity approach where we considered only high-fidelity (60 x 60 Ising) run, with 4 initial samples and next 4 sampling through BO. In other words, we did not consider any information from low-fidelity (20 x 20 Ising) run. For the purposes of validation, we considered 8 high-fidelity runs which took similar runtime (cost) with our full iMFBO analysis as in Fig. 7. To avoid bias in the result, we have considered 5 different sets of initial samples randomly realized. From **Fig. 10**, we can clearly see, the total number of runs is not sufficient to converge as the high-fidelity function prediction underfits. From the earlier iMFBO analysis, we know that the heat capacity function generated from high and low fidelity models are similar at low values and the deviation increases at high values, near the phase transition. However, in the classical single fidelity BO, we see significant deviation of high-fidelity function mean (blue line) from the low-fidelity function (black dotted line), even at the low heat capacity region. We suspect that this is due to the reason of not considering valuable information of cheaper low-fidelity data, along with high fidelity observations. Given that, in both the single and multifidelity (BO vs iMFBO) the cost of analysis is similar, iMFBO accelerates the learning with information gain from low-fidelity data, whereas the classical BO still needs more observations to reach that level of learning. Finally, we can see the MSEs (with mean MSE of $1.6 \times 10^{-3}$ and standard deviation of $0.89 \times 10^{-3}$) of single fidelity runs are overall much higher than most of the multi-fidelity runs. The individual SE maps are provided in the Supplementary **Fig. B4.** Table 1 summarizes the performance of all the single and multi-fidelity runs, where the interactive iMFBO, considering the correct low-fidelity model, have the best performance with 81% improvement in minimizing the MSE between ground truth and high fidelity predicted function than the average performance of single fidelity runs. When considered the incorrect low-fidelity model, the performance however reduced to 37.5% but still better than single fidelity run. For the non-interactive architectures, the sMFBO (data + physics driven) surpassed convincingly than the classical MFBO (data driven).

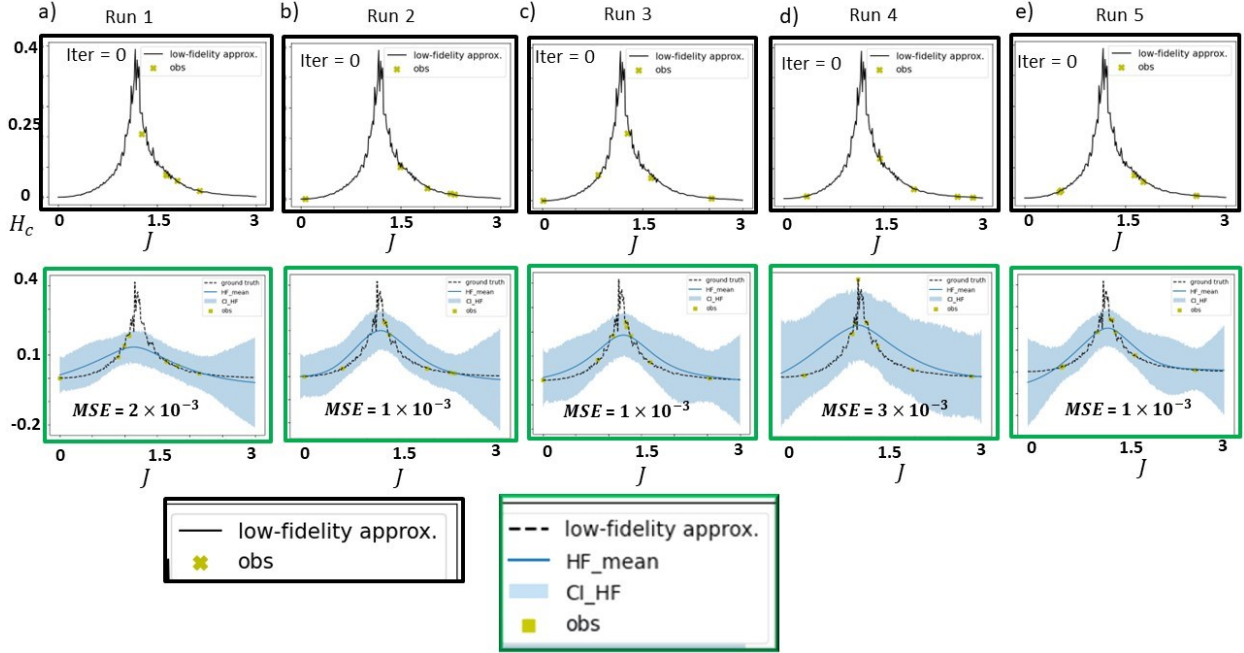

**Fig 10.** Classical Single-fidelity BO exploration over Ising models where the input, output and the fidelities of the model are same as mentioned in Fig. 5. Here, the black line is the referenced function of heat capacity of 20 x 20 Ising model (low fidelity). It is to be noted, low-fidelity data is not considered in this case and the BO only focused on high fidelity sampling of 60 x 60 Ising model denoted by yellow squares. We run 4 BO iterations which possess a similar cost of analysis in iMFBO as shown in fig 7. Figs (a)-(e) are the BO runs with 5 different random initial sampling. The top figure shows the initial sampled data (denoted by yellow cross), and the bottom figure shows the heat capacity functional form prediction (blue line) after 4 BO runs. The legends highlighted in black and green are the zoomed legends in the top and bottom figures respectively.

**Table 1.** Summary of the performance of different single fidelity and multi-fidelity BO.

| Algorithm | | MSE | %Improvement |
|---|---|---|---|
| BO (Single fidelity) | Run 1 | $2 \times 10^{-3}$ | NA |
| | Run 2 | $1 \times 10^{-3}$ | NA |
| | Run 3 | $1 \times 10^{-3}$ | NA |
| | Run 4 | $3 \times 10^{-3}$ | NA |
| | Run 5 | $1 \times 10^{-3}$ | NA |
| MFBO | HF: Square Ising model (60x60) LF: Square Ising model (20x20) | $0.8 \times 10^{-3}$ | 50% |
| | HF: Square Ising model (60x60) LF: Triangular Ising model (20x20) | $4 \times 10^{-3}$ | $-150\%$ |
| sMFBO (This work) | HF: Square Ising model (60x60) LF: Square Ising model (20x20) | $0.7 \times 10^{-3}$ | 56% |

|  | HF: Square Ising model (60x60) | $0.5 \times 10^{-3}$ | 68% |
|  | LF: Triangular Ising model (20x20) | | |
| iMFBO (This work) | HF: Square Ising model (60x60) | $0.3 \times 10^{-3}$ | 81% |
|  | LF: Square Ising model (20x20) | | |
|  | HF: Square Ising model (60x60) | $1 \times 10^{-3}$ | 37.5% |
|  | LF: Triangular Ising model (20x20) | | |

**Note:** The percentage of improvement is calculated from the mean MSE of the single-fidelity BO run. The percentage of improvement cell for MFBO-Triangular Ising model is negative as it worsen the performance than single fidelity run.

### 2.3. Current Challenges and Future Opportunities:

To summarize, through exploration over the Ising model with non-interactive (data and/or physics driven) and interactive MFBO architecture (combining data, physics, and human real time decision driven), here we have summarized a few key challenges. It is evident to say that though the Ising model is well established, however for the development of BO algorithm in the multi-fidelity setting, it showed some non-trivial characteristics which are hard to solve in an optimization problem and can be easily think of many real world physics and material science problems. We observed that due to the non-smoothness of the multi-fidelity problem, the search process requires a longer time and over usage to first learn the low fidelity function itself, before gaining precision knowledge with high-fidelity explorations. Also, having a high degree of non-smoothness at the region of interest, we see the exploitation of the good region worsens the overall performance of the model, where the model uncertainty increases significantly throughout and required additional exploration to reduce the uncertainty. We have shown to gather better insights of the high-fidelity functional form with interactive policy changes and inclusion of prior physical knowledge.

Through exploring the physics based Ising model via different multi-fidelity Bayesian optimization, which not only gave the scope to learn 60x60 Ising model in reasonable time, but also provides potential room for improvement in algorithm development of human in the loop based automated experiments (AE) for other complex problems towards material discovery, which is the main purpose of this study. As potential future opportunities, we aim to include other policies which can be operated interactively as needed through monitoring and on the fly knowledge from exploration. For example, the exploration can be better strategies with physics driven multi-fidelity acquisition function which can trigger where and how much exploration should be aimed or avoided in order to increase the overall performance of the MFGP fit. Secondly, we can also include the policy of adaptively (suggested by ML process) and selectively (suggested by human) processing the data on the fly (eg. smoothing the data) to accelerate the search without overshooting it. Thirdly, we can define a policy to segment the search space based on the level of complexities as an example to use multiple GPs – one to fit the low complex smooth region and another to fit high complex non-smooth regions.

**Conclusion**:

We have proposed a human in the loop workflow for physics discovery based on multifidelity Bayesian optimization and structured Gaussian processes. We observe that pure data driven multideity processes can accelerate the initial discovery. However, the reliance on the symmetric data-driven fidelity kernel slows down the discovery in the later stages. The physics-driven model based on structured GP allows to incorporate partially known physics in the form of probabilistic model. In these settings, both this physical knowledge and low-fidelity data are used to accelerate the high-fidelity exploration.

We propose and realize the human-in-the-loop interactive iMFBO workflow that allows balancing the MF and physics-driven explorations, highlight current challenges, and identifying potential opportunities for algorithm development that combines data, physics, and real-time human decisions.

**Method:**

The multi-fidelity co-variance function is defined as

$$cov(x^i, x^j, f^i, f^j) = \sigma^2 \times R(x^i, x^j) \times K_F(f^i, f^j) \tag{5}$$

$$R(x^i, x^j) = \exp\left(-0.5 \times \sum_{m=1}^{d} \frac{(x_m^i - x_m^j)^2}{\theta_m^2}\right) \tag{6}$$

$$R(x^i, x^j) = \exp\left(-\sqrt{5} \times \sum_{m=1}^{d} \frac{|x_m^i - x_m^j|}{\theta_m}\right) \times \left(1 + \sqrt{5} \times \sum_{m=1}^{d} \frac{|x_m^i - x_m^j|}{\theta_m} + \frac{5}{3} \times \sum_{m=1}^{d} \frac{|x_m^i - x_m^j|^2}{\theta_m^2}\right) \tag{7}$$

$$K_F(x^i, x^j) = \exp(-\delta|f^i - f^j|) \tag{8}$$

where $\sigma^2$ is the overall variance parameter, $\theta_m$ is the correlation length scale parameter in dimension $m$ of $d$ dimension of $x$, $\delta > 0$ is the learning parameter which measures the fidelity gap between the inputs. These are termed as the hyper-parameters of $\Delta^F$. $R(x^i, x^j)$ is the spatial correlation function in the input data space which is considered as radial basis function (eq. 6) or matern kernel function (eq. 7), $K_F(f^i, f^j)$ is the correlation function in the fidelity space.

The multi-fidelity acquisition function is defined as

a. Calculate the high-fidelity acquisition values as

$$EI(\bar{\bar{y}}(\bar{\bar{x}})|f = 1)) =$$

$$\begin{cases} (\mu_{HF}(\bar{\bar{y}}(\bar{\bar{x}})) - y(x^+) - \xi) \times \Phi(Z, 0, 1) + \sigma_{HF}(\bar{\bar{y}}(\bar{\bar{x}})) \times \phi(Z_{HF}) & if\ \sigma_{HF}(\bar{\bar{y}}(\bar{\bar{x}})) > 0 \\ 0\ if\ \sigma_{HF}(\bar{\bar{y}}(\bar{\bar{x}})) = 0 \end{cases} \tag{9}$$

$$Z_{HF} = \begin{cases} \frac{\mu_{HF}(\bar{\bar{y}}(\bar{x})) - y(x^+) - \xi}{\sigma_{HF}(\bar{\bar{y}}(\bar{x}))} & if\ \sigma_{HF}(\bar{\bar{y}}(\bar{x})) > 0 \\ 0 & if\ \sigma_{HF}(\bar{\bar{y}}(\bar{x})) = 0 \end{cases} \qquad (10)$$

b. Calculate the low-fidelity acquisition values as

$$\Delta EI(\bar{\bar{y}}(\bar{x})) = |EI(\bar{\bar{y}}(\bar{x}|f=1)) - EI(\bar{\bar{y}}(\bar{x}|f=0))| \qquad (11)$$

$$EI(\bar{\bar{y}}(\bar{x}|f=0)) =$$
$$\begin{cases} (\mu_{lf}(\bar{\bar{y}}(\bar{x})) - y(x^+) - \xi) \times \Phi(Z,0,1) + \sigma_{lf}(\bar{\bar{y}}(\bar{x})) \times \phi(Z_{lf}) & if\ \sigma_{lf}(\bar{\bar{y}}(\bar{x})) > 0 \\ 0 & if\ \sigma_{lf}(\bar{\bar{y}}(\bar{x})) = 0 \end{cases}$$
$$(12)$$

$$Z_{lf} = \begin{cases} \frac{\mu_{lf}(\bar{\bar{y}}(\bar{x})) - y(x^+) - \xi}{\sigma_{lf}(\bar{\bar{y}}(\bar{x}))} & if\ \sigma_{lf}(\bar{\bar{y}}(\bar{x})) > 0 \\ 0 & if\ \sigma_{lf}(\bar{\bar{y}}(\bar{x})) = 0 \end{cases} \qquad (13)$$

c. Calculate the multi-fidelity acquisition values as

$$u(x,f) = \begin{cases} \frac{EI(\bar{\bar{y}}(\bar{x}|f=1))}{C} & if\ f = 1 \\ \Delta EI(\bar{\bar{y}}(\bar{x})) & if\ f = 0 \end{cases} \qquad (14)$$

d. Maximize the multi-fidelity acquisition function at BO iteration k

$$\max_{x_{k+1}, f_{k+1}} U(\overline{X_k^F}) \qquad (15)$$

where $y(x^+)$ is the current maximum value among all the sampled data in both fidelity spaces until the current stage, i.e. in $X_k^F$; $\mu_{HF}(\bar{\bar{y}})$, $\mu_{lf}(\bar{\bar{y}})$ and $\sigma_{HF}^2(\bar{\bar{y}})$, $\sigma_{lf}^2(\bar{\bar{y}})$ are the predicted high-fidelity and low-fidelity mean and standard deviation from MFGP; $\Phi(.)$ is the cdf; $\phi(.)$ is the pdf; $\xi \geq 0$ is a small value which is set as 0.01, $C$ is the cost ratio between high and low fidelity models which can be derived from model computational complexities and domain knowledge.

**Additional Information:**

See the supplementary material for additional analysis and figures, related to the research.

**Acknowledgements:**

This work (A.B) was supported by the US Department of Energy, Office of Science, Office of Basic Energy Sciences, MLExchange Project, award number 107514, and supported (M.V., S.V.K.) by the US Department of Energy, Office of Science, Office of Basic Energy Sciences, as part of the Energy Frontier Research Centers program: CSSAS–The Center for the Science of Synthesis Across Scales–under Award Number DE-SC0019288. The Ising model research (R.V) was also supported by the Center for Nanophase Materials Sciences (CNMS), which is a US Department of Energy, Office of Science User Facility at Oak Ridge National Laboratory. MZ acknowledges the Laboratory Directed Research and Development Program at Pacific Northwest National Laboratory, a multiprogram national laboratory operated by Battelle for the U.S. Department of Energy.

**Conflict of Interest:**

The authors declare no conflict of interest.

**Code and Data Availability Statement:**

The analysis reported here along with the code is summarized in Colab Notebook for the purpose of tutorial and application to other data and can be found in https://github.com/arpanbiswas52/iMFBO_Ising

Supplementary Materials of the paper titled

**Towards accelerating physical discovery via non-interactive and interactive multi-fidelity Bayesian Optimization: Current challenges and future opportunities**

Arpan Biswas[1,a], Sai Mani Prudhvi Valleti[2], Rama Vasudevan[3],
Maxim Ziatdinov,[4] Sergei V. Kalinin[4,5,b]

[1] University of Tennessee-Oak Ridge Innovation Institute, Knoxville, USA
[2] Bredesen Center for Interdisciplinary Research, University of Tennessee, Knoxville, USA
[3] Center for Nanophase Materials Sciences, Oak Ridge National Laboratory, USA
[4]Physical Sciences Division, Pacific Northwest National Laboratory, Richland, WA 99352 USA

[5]Department of Materials Science and Engineering, University of Tennessee, Knoxville, USA

## Appendix A. Test Problems

### Problem 1.

$$F_1 = -(x+1)^2 \times \frac{\sin(2x+2)}{5} + 1 + \frac{x}{3} \tag{S1}$$

$$f_1 = \frac{F_1}{2} + \frac{x}{4} + 2 \tag{S2}$$

### Problem 2.

$$F_2 = \begin{cases} F_1 \ if \ x < 7.5 \\ F_1 - 15 \ if \ x \geq 7.5 \end{cases} \tag{S3}$$

$$f_2 = \frac{F_1}{2} + \frac{x}{4} - 5 \tag{S4}$$

where $F$ denotes high-fidelity function and $f$ denotes the low-fidelity function of the respective test problems.

**Appendix B. Additional figures**

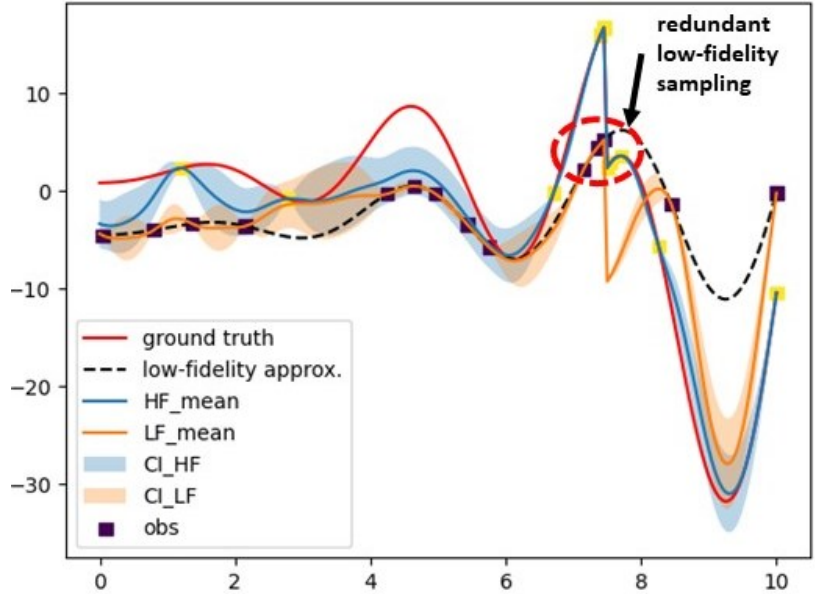

**Figure B1.** Additional figure of the case study in Fig 3a-b, where after finding the discontinuity in the high-fidelity model, the non-interactive BO suggests redundant low-fidelity sampling at that region (circled in red).

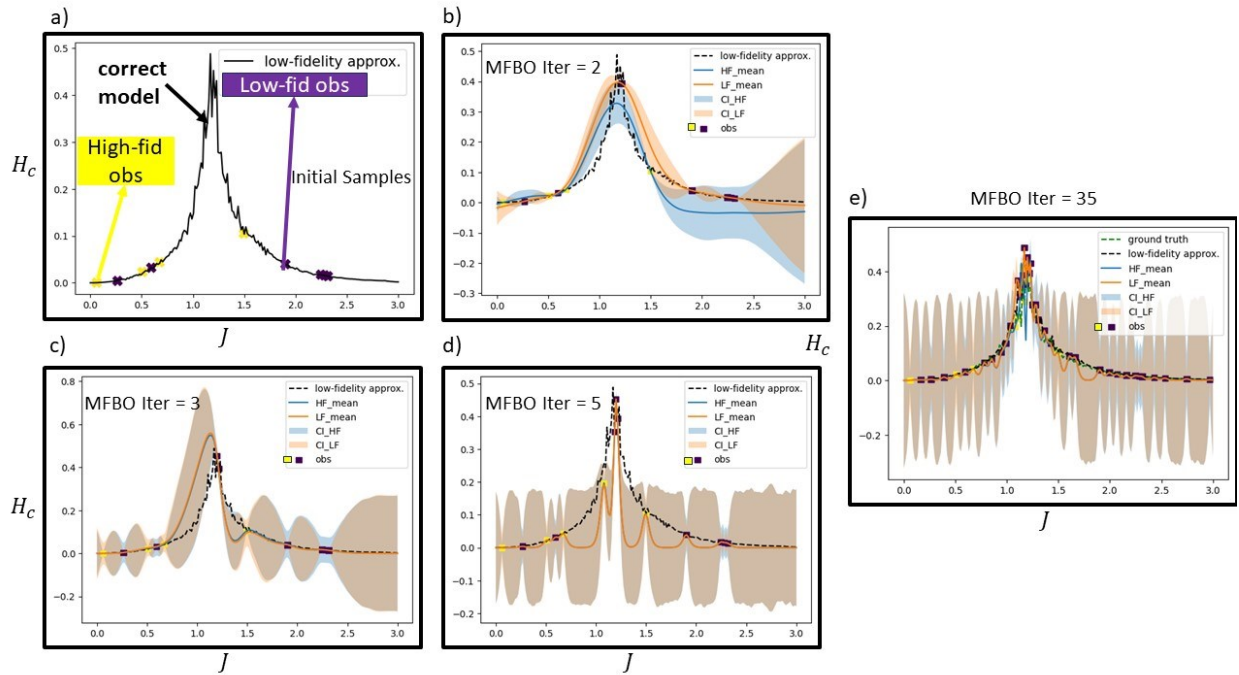

**Fig B2.** Non-interactive standard Multi-fidelity BO (MFBO) exploration over Ising models where the input, output and the fidelities of the model are same as mentioned in Fig. 5. Here, we consider

the correct model of interest (square model) in the low-fidelity space (denoted by black line) in fig a. The yellow and the black squares in the legend of fig. b- e denote the high-fidelity and low-fidelity observations, as autonomously sampled from MFBO.

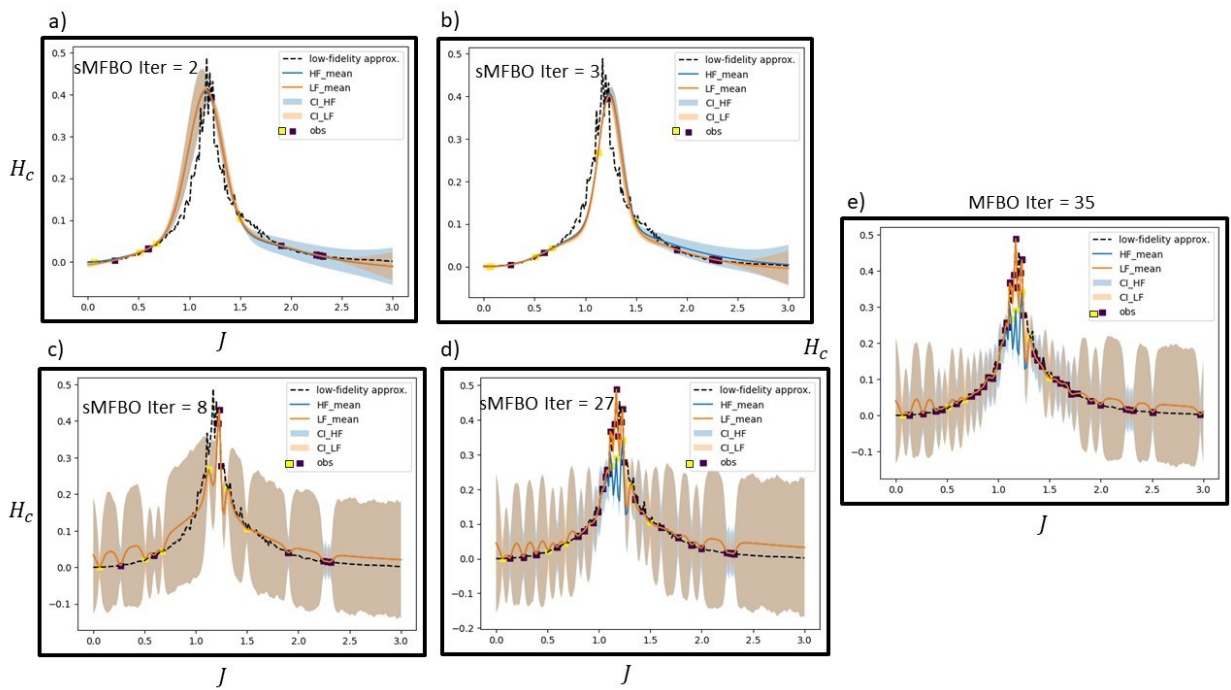

**Fig B3.** Non-interactive structured Multi-fidelity BO (sMFBO) exploration over Ising models where the input, output and the fidelities of the model are same as mentioned in Fig. 5. Here, we consider the correct model of interest (square model) in the low-fidelity space (denoted by black line) in fig a. The yellow and the black squares in the legend of the figures denote the high-fidelity and low-fidelity observations, as autonomously sampled from sMFBO.

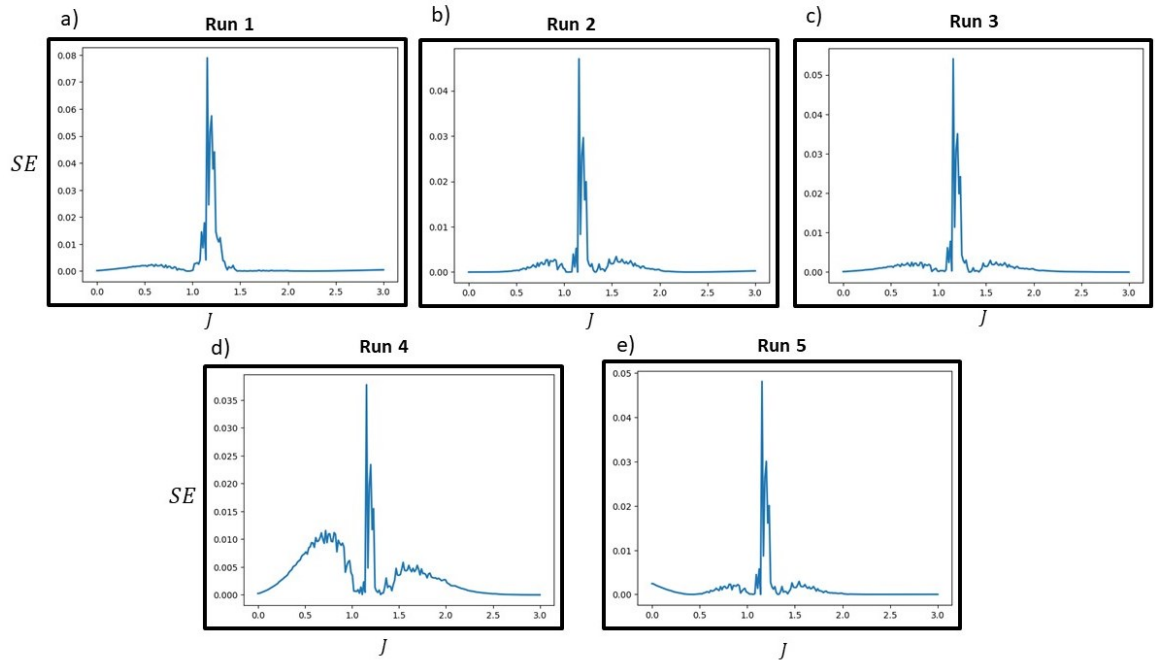

**Fig B4.** Square errors from Classical Single-fidelity BO exploration over Ising models where the input, output and the fidelities of the model are same as mentioned in Fig. 5 where the BO runs with 5 different random initial sampling, and then with 4 BO iterations as referred in Fig 10.